\documentclass[sigconf,nonacm]{acmart}
\AtBeginDocument{%
  }

\usepackage{multirow}
\usepackage{bm}

\usepackage{amsmath,amssymb,amsthm}
\newcommand{\R}{\mathbb{R}}
\usepackage{array}
\usepackage{tabularx}
\usepackage{enumitem}

\newcommand{\bq}{\mathbf{q}}
\newcommand{\ba}{\mathbf{a}}
\newcommand{\bt}{\mathbf{t}}

\begin{document}

\title{LACRM: LLM-Guided Mining of Directed Advertiser Competition Relationships in E-Commerce Advertising}

\author{Xiang Shi$^{1*}$, Yitian Kou$^{1,2*}$, Bing Li$^{1\dagger}$, Shengsheng Niu$^{1\dagger}$, Chun Gan$^{1}$, Linghui Meng$^{1}$,\\
Jie He$^{1}$, Zhangang Lin$^{1}$ and Ching Law$^{1}$}
\affiliation{%
  \institution{$^{1}$JD.com, Beijing, China \qquad $^{2}$East China Normal University, Shanghai, China}
  \country{}}
\email{\{shixiang12, libing200, niushengsheng, ganchun1, menglinghui1, hejie67, linzhangang, lawching\}@jd.com, yitian.kou@stu.ecnu.edu.cn}
\renewcommand{\shortauthors}{Shi et al.}


\begin{abstract}
In e-commerce advertising, ad platforms typically fail to model the externalities among advertisers when making optimization decisions: an advertiser's actual performance is largely shaped by rivals competing for the same users; these rivals typically operate in overlapping product categories and at comparable budget scales. However, capturing this rivalry as an explicit, reusable signal is challenging because it is directed, multi-faceted, and partially observable. Existing paradigms such as interaction graphs, taxonomy rules, and single-embedding models do not fully capture these properties.
In this paper, we propose \textbf{LACRM} (\textbf{L}LM-driven \textbf{A}dvertiser \textbf{C}ompetition \textbf{R}elationship \textbf{M}ining), a novel framework that constructs directed competition graphs with continuous edge intensities via a three-stage pipeline.
LACRM moves beyond symmetric similarity through LLM-driven role-separated profiling, which enables asymmetric multi-route retrieval to uncover both realized and latent rivals. The framework further incorporates listwise learning-to-rank aligned with temporal auction behavior to accurately model directed competition intensity.
In offline retrieval experiments on an industrial benchmark, LACRM outperforms strong directed GNN baselines and ablation variants. Multi-LLM and human studies provide complementary evidence that the recovered latent rivals reflect meaningful market competition rather than merely model-generated artifacts. In production A/B tests, LACRM-guided bidding and incentive allocation increase advertising revenue by 1.73\% and 1.14\%, respectively.
The results demonstrate the practical value of explicit competition modeling in large-scale advertising systems.
\end{abstract}



\maketitle

{\renewcommand{\thefootnote}{\fnsymbol{footnote}}%
\footnotetext[1]{These authors contributed equally to this work.}%
\footnotetext[2]{Corresponding authors.}}%

\section{Introduction}
\label{sec:introduction}

Modern e-commerce advertising platforms coordinate three tightly coupled stakeholders: users who generate traffic demand, advertisers who compete for exposure, and the platform that allocates ad opportunities via auction and delivery mechanisms. Existing advertiser-facing intelligence systems support automated bidding \cite{10.1145/3018661.3018702,10.1145/3269206.3271748}, campaign delivery and budget pacing \cite{10.1145/2783258.2788615, 10.1145/2623330.2623366}, and merchant incentives \cite{liu2019merchant}, but typically optimize each advertiser as an independent decision unit. Their predictions rely primarily on the advertiser's own attributes, campaign history, and bidding outcomes. Such pointwise modeling neglects an important source of market context: an advertiser's performance depends not only on its own actions, but also on the competitors that target overlapping users, products, and budgets \cite{edelman2007gsp, VARIAN20071163}. Indeed, advertiser-to-advertiser competition is rarely represented as an explicit, reusable signal in current advertiser-facing systems; instead, it is only inferred indirectly from realized auction outcomes.

An explicit competition layer provides relational context across these advertiser-facing decisions. More broadly, a competition graph offers a shared market representation that many downstream systems can consume, rather than requiring each application to infer competition independently from its own behavioral logs.

However, constructing such a competition graph presents non-trivial challenges along three critical dimensions. First, advertiser competition is inherently directed. A dominant flagship store can heavily siphon traffic from a niche merchant, whereas the reverse impact is negligible. Empirically, 99.24\% of the 70 million observed competition edges show marked asymmetry in direction or intensity (Sec.~\ref{sec:problem}).
Second, competition manifests across multiple facets. Advertisers interact through direct category and price overlaps~\cite{mcauley2015substitutable}, intra-brand channel friction~\cite{chiang2003direct,tsay2004channel}, or cross-category substitution~\cite{ratneshwar1996goal,ratneshwar1991substitution} driven by shared consumer needs. As a result, a single similarity representation can be dominated by frequent, homogeneous interactions, masking less frequent but distinct competition patterns.
Third, behavioral evidence is inherently incomplete. While auction co-occurrence and interaction graphs record realized competition, they provide no direct evidence for latent competitors that have yet to participate in shared auctions. This limitation is particularly acute for new and long-tail advertisers.

Existing relation-mining approaches do not fully address these challenges. Taxonomy- and rule-based approaches ensure broad coverage yet lack the capacity to express fine-grained competition strength. While behavioral graph models (e.g., GNN-based link prediction) can encode directed graph structures, their predictions remain constrained by the limited coverage of observed interaction graphs and historical edges \cite{hibshman2023inherentlimitstopologybasedlink}.
Semantic representations provide a complementary source of evidence. Specifically, Large Language Models (LLMs) can consolidate heterogeneous merchant attributes into structured semantic profiles and reason about functional utility, target consumer intents, price positioning, and shopping scenarios. This open-world knowledge enables the retrieval of plausible competitors beyond historical auction interactions.
However, directly applying a shared text embedding still conflates the distinct roles of the advertiser being modeled and the competitor being retrieved, and therefore resolves neither directionality nor multi-faceted competition. Dedicated competitor-mining methods operate at the firm or product level and typically model competition as symmetric, undirected, or binary pairwise relations~\cite{chen2020studying,ding2024jpec,intercorprel2025}. Although asymmetric competitive effects have long been studied at the brand level in marketing science~\cite{carpenter1988modeling}, modeling directed and intensity-aware competition at the advertiser level remains underexplored, and, to our knowledge, no dedicated public benchmark is currently available.

In this paper, we formulate advertiser competition relationship mining as the problem of inferring a directed, weighted graph that quantifies asymmetric competitive intensities among advertisers. To address this challenge, we propose \textbf{LACRM} (\textbf{L}LM-driven \textbf{A}dvertiser \textbf{C}ompetition \textbf{R}elationship \textbf{M}ining), a framework for constructing directed advertiser competition graphs with continuous edge weights. LACRM represents each advertiser through an anchor-side Advertiser Representation that captures the merchant's market identity and multiple query-side representations that specify relevant competitors across distinct competition views. It then performs asymmetric multi-route retrieval to surface both realized and latent competitors, followed by a listwise learning-to-rank module that calibrates semantic candidates against temporal auction behaviors to estimate relative competition intensities. This design combines the broad coverage of LLM-derived semantic representations with the precision of observed behavioral evidence.

The contributions in this paper can be summarized as follows:

(i) We formulate advertiser-level competition mining in e-commerce advertising as a directed, weighted edge-ranking problem and introduce an explicit competition layer that provides a reusable market representation for advertiser-facing optimization tasks.

(ii) We propose LACRM, which combines role-separated advertiser and query profiles, multi-route semantic retrieval, and listwise behavioral alignment to capture directed competition and recover latent rivals beyond observed auction graphs.

(iii) We evaluate LACRM on industrial e-commerce advertising data through realized-competitor retrieval, multi-model assessment of latent competitors corroborated by an independent human expert study, and two production applications. LACRM outperforms its ablated variants and two directed GNN baselines, while its deployment in downstream bidding and incentive allocation yields measurable gains in advertising revenue and budget fulfillment.

\section{Problem Formulation}
\label{sec:problem}
We formalize the implicit market competition structure as a directed, weighted competition graph $G=(V,E,W)$, where nodes $V=\{v_1,\dots,v_N\}$ denote advertisers (used interchangeably with merchants) associated with feature profiles $\{\mathbf{x}_i\}_{i=1}^N$. A directed edge $e_{ij} \in E$ signifies that advertiser $v_i$ regards $v_j$ as a competitor, with its directional weight formulated as:
\begin{equation}
w_{ij} = s(v_i \to v_j) \in [0, 1]
\end{equation}
This score quantifies the intensity of competitive pressure exerted by $v_j$ from the asymmetric perspective of $v_i$. Our objective is to learn the scoring function $s(\cdot)$ to construct the dense competitive strength matrix $\mathbf{W} \in \mathbb{R}^{N \times N}$, which serves as a foundational market-structure representation for downstream tasks.

To guide the learning of this graph, our framework is grounded in two core principles:
\begin{itemize}
    \item \textbf{Continuous Intensity}: Competition is characterized by the intensity of traffic diversion from $v_i$ attributable to $v_j$, operationally approximated using auction co-occurrence signals.
    \item \textbf{Latent Discovery}: Auction co-occurrence provides a reliable yet incomplete signal of realized rivalry, necessitating the recovery of latent competitors beyond historical interactions.
\end{itemize}

The competition score is inherently asymmetric ($s(v_i\!\to\! v_j) \neq s(v_j\!\to\! v_i)$). Across a corpus of over 70 million directed competition edges, 77.41\% are strictly unidirectional, lacking a reciprocal edge. Among the remaining reciprocated edges, 96.62\% exhibit distinct competitive intensities. In total, only 0.76\% of all edges are fully symmetric with equal intensity in both directions, leaving 99.24\% asymmetric in either direction or intensity. For instance, consider a niche accessory shop $v_B$ and a dominant head-brand flagship store $v_A$ operating within the same category. While $v_B$ faces severe suppression from $v_A$ (yielding a high $s(v_B\!\to\! v_A)$), the market leader $v_A$ remains largely indifferent to $v_B$ ($s(v_A\!\to\! v_B)\approx 0$).

\begin{figure*}[t]
\centering

\includegraphics[width=\textwidth]{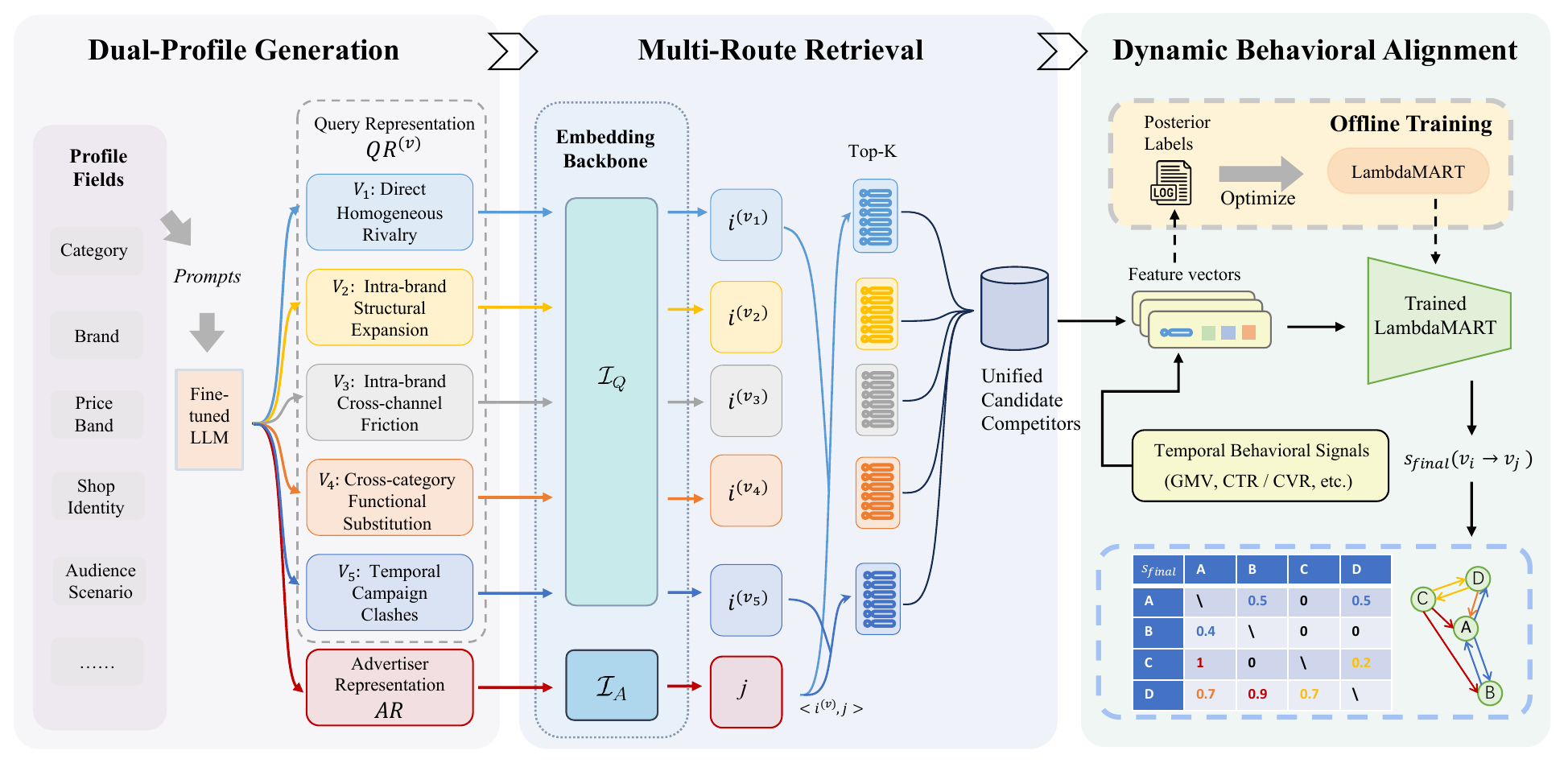}
\caption{System architecture of LACRM. Following the pipeline flow ($\mathbf{x}_i \to \text{Profiles} \to \text{Candidates} \to \text{Score}$), Module~1 generates the isomorphic Advertiser Representation and multi-view Query Representation; Module~2 performs instruction-conditioned asymmetric embedding and multi-route retrieval; and Module~3 reranks the retrieved candidates using temporal behavioral signals to produce the final directed competition scores.}
\label{fig:framework}
\end{figure*}

\section{Methodology}
As illustrated in Figure~\ref{fig:framework}, our framework converts structured merchant profiles into continuous asymmetric competition intensities across three sequential stages: Dual-Profile Generation, Multi-Route Retrieval, and Dynamic Behavioral Alignment.

\subsection{LLM-Guided Dual-Profile Generation}
\label{sec:profile}

To represent directional commercial rivalry without collapsing the proximity matrix into a symmetric similarity space, we structurally decouple each advertiser's semantic footprint into an asymmetric dual-view representation.
We define the LLM-guided \textbf{Advertiser Representation (AR)} and \textbf{Query Representation (QR)} to respectively formulate the static identity and dynamic intent of a merchant. AR is a factual profile that captures "who the merchant is" to serve as the invariant retrieval anchor, whereas QR characterizes a strategic, multi-view trajectory identifying "which cohort of rivals the merchant targets" to serve as the variable retrieval query.

To keep the two roles structurally aligned before embedding, we constrain them to an isomorphic six-field schema $\mathcal{F}=\{f_1,\dots,f_6\}$: \textit{Category \& Brand} ($f_1$, L3 category hierarchy and brand-camp identity), \textit{Search Intent} ($f_2$, core queries and keyword distribution), \textit{Style \& Positioning} ($f_3$, design language and market tier), \textit{Audience \& Scenario} ($f_4$, consumer personas and usage scenarios), \textit{Price Band} ($f_5$, dominant per-item price range in CNY), and \textit{Shop Identity} ($f_6$, distribution channel and GMV tier). Throughout, we follow the platform's product-taxonomy notation, in which \emph{L3} denotes a fine-grained, near-leaf (third-level) category and \emph{L2} its coarser parent level. The target AR is consistently single-view and fully populated by concatenating all six fields, $\mathbf{AR}_i = \bigoplus_{k=1}^{6} f_k(\mathbf{x}_i)$.

In contrast, a single query vector conflates all competition types. High-frequency category overlaps then dominate retrieval, systematically suppressing subtler but strategically important rivalries~\cite{li2019multi, chen2020studying}. To enforce relational separation and provide dedicated retrieval capacity for less frequent competition patterns, we decompose the query space into a taxonomy of five fine-grained competitive views ($\mathcal{V}$).

Rather than positing an ad-hoc list, we derive these views from a principled partition of the competitive space along two fundamental axes: brand relation (inter- vs.\ intra-brand) and category relation (same L3 fine-grained category vs.\ cross-category), augmented by a temporal overlay. This operationalizes the classical marketing framework: competition is stratified by substitutability and structured hierarchically rather than flatly, spanning from head-to-head brand rivalry through category-level competition to cross-category functional substitution~\cite{kotler2016marketing, urban1984testing}. 

The four structural views explicitly capture the distinct quadrants of this brand-category space. Specifically, the overlapping intra-brand and same-category cell is further refined via the shop-identity channel axis ($f_6$) into \textit{Intra-Brand Cross-Channel Friction}, while \textit{Temporal Campaign Clashes} operates as an independent, gated route overlaid across the space. Each view is instantiated purely through view-conditioned field masking coupled with a dedicated retrieval route, establishing a strict one-to-one mapping between competition views and retrieval routes. We formalize each view's competitive boundary and active field subset in Appendix~\ref{app:views}.


For each view $\omega\in\mathcal{V}$ with active fields
$\mathcal{F}^{(\omega)}\subseteq\mathcal{F}$, we define

\begin{equation}
\mathbf{QR}_i^{(\omega)} = \tau(\omega) \,\Vert\, \bigoplus_{k=1}^{6} \psi_\omega\big(f_k(\mathbf{x}_i)\big)
\end{equation}
where $\tau(\omega)$ denotes a view-specific instruction prefix, and $\psi_\omega$ applies field masking that retains only the fields in $\mathcal{F}^{(\omega)}$ while setting unselected fields to empty tokens.

Profile synthesis is grounded in each advertiser's realized competitive history when available. For an advertiser with observed rivals, its stable realized competitors from a strictly prior three-month window are supplied to the generator as few-shot anchors that seed and diversify the query views; the model is instructed to surface rivals beyond these anchors rather than restate them, so the anchors bias recall toward the correct competitive neighborhood without bounding it to already-observed pairs. For cold-start advertisers with no realized competitors, this field is empty and generation reduces to domain-knowledge-driven inference based solely on the merchant's own attributes.

Profile generation is trained primarily with supervised fine-tuning (SFT) on \texttt{Qwen3-\allowbreak{}4B-\allowbreak{}Instruct}~\cite{qwen3technicalreport} using AR and multi-view QR data from the \texttt{GPT-5.5} teacher, followed by auxiliary Group Relative Policy Optimization (GRPO)~\cite{shao2024deepseekmathpushinglimitsmathematical}. We enforce format compliance as a hard gate:
\begin{equation}
R(\mathbf{QR},\omega)=
\begin{cases}
R_{\text{format}}, & \text{invalid/degenerate},\\
r_{\text{inactive}}, & \text{correct inactive exit},\\
\mathrm{clip}(\rho_rR_{\text{retr}}+\rho_jR_{\text{judge}},r_{\min},r_{\max}), & \text{otherwise}.
\end{cases}
\end{equation}
Invalid outputs receive deterministic negative penalties and bypass further scoring. For valid active views, $R_{\text{retr}}$ measures view-conditioned retrieval quality against held-out historical realized rivals, with reward labels separated from prompt anchors, ranker labels, and final evaluation; $R_{\text{judge}}$ averages heterogeneous training-time LLM judges. Historical positives are filtered by the view boundary, with judge-only fallback when none exist. We set $\rho_r=0.6$ and $\rho_j=0.4$ to place greater weight on retrieval quality, while retaining the frozen SFT model as the KL reference. Full details are provided in the supplementary material.

The resulting model is subsequently used as the profile-generation LLM throughout the LACRM pipeline.

\subsection{Asymmetric Embedding and Multi-Route Retrieval}
\label{sec:asymmetric_embedding}

Following the recent success of instruction-conditioned representation paradigms~\cite{su2023one}, we model the required asymmetry through task-differentiated prompts on top of the pretrained \texttt{Qwen3-Embedding-8B}~\cite{qwen3embedding} backbone, without any additional embedding-model training. To this end, the generative Query Representation $\mathbf{QR}_i^{(\omega)}$ and static Advertiser Representation $\mathbf{AR}_j$ are paired with asymmetric task instructions $\mathcal{I}_Q$ and $\mathcal{I}_A$. The resulting $\ell_2$-normalized dense query embedding $\bq_i^{(\omega)}$ and target anchor embedding $\ba_j$ are extracted as follows:
\begin{equation}
\bq_i^{(\omega)} = \Phi\big(\mathbf{QR}_i^{(\omega)}, \mathcal{I}_Q\big), \qquad \ba_j = \Phi\big(\mathbf{AR}_j, \mathcal{I}_A\big)
\end{equation}
where $\Phi(\cdot, \cdot)$ denotes the embedding function that maps a text--instruction pair into the continuous representation space.

The base semantic competition score under a designated view $\omega$ is subsequently evaluated via the inner product $s_{\text{emb}}^{(\omega)}(v_i\!\rightarrow\! v_j)=\big\langle \bq_i^{(\omega)},\ \ba_j\big\rangle$. Because the two directions use role-separated input texts and task-specific instructions, this defines an asymmetric scoring function: $s_{\text{emb}}^{(\omega)}(v_i\!\rightarrow\! v_j)$ is not constrained to equal $s_{\text{emb}}^{(\omega)}(v_j\!\rightarrow\! v_i)$.

To avoid conflating heterogeneous competition types in a single retrieval pass, we run an independent retrieval route per view. For each view $\omega$, the system queries the global anchor index to fetch the Top-$k_r$ nearest neighbors (with $k_r=300$ per view in our experiments) based on directional affinity:
\begin{equation}
\mathcal{C}_\omega(v_i)=\operatorname{TopK}_{k_r}\!\Big(\big\{s_{\text{emb}}^{(\omega)}(v_i\!\rightarrow\! v_j)\big\}_{v_j\in V}\Big)
\end{equation}
The candidate edges retrieved from the view-specific routes are first
merged at the directed merchant-pair level, yielding the global
candidate set
\begin{equation}
\mathcal{C}(v_i)
=
\bigcup_{\omega\in\mathcal{V}} \mathcal{C}_\omega(v_i).
\end{equation}
Rather than collapsing the view-specific scores into a single
pre-aggregated value, we pivot them into a fixed $|\mathcal{V}|$-dimensional
semantic feature vector $\mathbf{s}_{\mathrm{emb}}(v_i\!\rightarrow\!v_j)\in\mathbb{R}^{|\mathcal{V}|}$
whose $\omega$-th entry is the view-$\omega$ score
$s_{\mathrm{emb}}^{(\omega)}(v_i\!\rightarrow\!v_j)$ if $v_j\in\mathcal{C}_\omega(v_i)$
and $0$ otherwise. With $|\mathcal{V}|=5$, each directed candidate pair carries
five view-specific scores, paired with a hit-indicator vector
$\mathbf{m}_{ij}=\big[\mathbb{I}(v_j\in\mathcal{C}_\omega(v_i))\big]_{\omega\in\mathcal{V}}$
that distinguishes an unobserved view from a genuine score of zero. Both are
passed to the downstream Dynamic Behavioral Alignment, letting the ranker learn
the views' relative importance and nonlinear interactions.

\subsection{Dynamic Behavioral Alignment}
\label{sec:rerank}

While instruction-conditioned embedding retrieval captures multi-view semantic commonalities, it does not directly encode immediate empirical market dynamics such as active auction collisions. To incorporate this evidence, we formulate the final refinement stage as a fine-grained listwise Learning-to-Rank (LTR)~\cite{liu2009learning} paradigm implemented with LambdaMART~\cite{burges2010ranknet}.

For each source advertiser $v_i$, the retrieved candidate edge set $\mathcal{C}(v_i)$ forms a ranking list. 
The final competition score is produced by a gradient-boosted decision tree ensemble:
\begin{equation}
s_{\text{final}}(v_i \rightarrow v_j) = \mathcal{H}_{\text{LTR}}(\mathbf{x}_{ij})
\end{equation}
where $\mathbf{x}_{ij}$ is a heterogeneous feature vector comprising the five-dimensional view-specific semantic score vector
$\mathbf{s}_{\mathrm{emb}}(v_i\!\rightarrow\!v_j)$, its corresponding hit-indicator vector $\mathbf{m}_{ij}$, and temporal behavioral
features derived from auction logs. LambdaMART therefore learns how the different competition views should be weighted and combined for
each candidate pair, together with their interactions with observed market behavior. Because LambdaMART produces unbounded ranking scores, we apply a monotonic per-source normalization that maps each candidate's raw score within a ranking list to $[0,1]$, yielding normalized competition weights that populate $\mathbf{W}$ (Sec.~\ref{sec:problem}) while preserving the learned ranking order.

To capture short-term operational fluctuations and real-time market friction, the raw temporal records of each merchant $v_i$ are structured into an $\ell_2$-normalized behavioral profile $\tilde\bt_i \in \R^{d_t}$ spanning six complementary groups of market dynamics: traffic scale, conversion efficiency, volumetric dynamics, ecosystem rhythm and lifecycle, behavioral quality, and intra-category quantiles and trends. For a directed candidate pair $(v_i, v_j)$, the temporal features feeding the ranker $\mathcal{H}_{\text{LTR}}$ are formed by concatenating the source and target profiles $\tilde\bt_i$ and $\tilde\bt_j$ with their pairwise interaction terms; the complete per-group feature construction is detailed in Appendix~\ref{app:features}. In the cold-start regime, $\mathbf{x}_{ij}$ contains only semantic and static features, allowing new market entrants to be scored even when behavioral features are unavailable.

\section{Experiments}
\label{sec:experiments}
We evaluate LACRM along three dimensions: (1) offline candidate retrieval against realized rivalries, (2) verification of latent rivalries, and (3) online A/B testing in a production advertising system.

\subsection{Experimental Setup}

\subsubsection{Dataset and Realized Ground Truth}
Our offline evaluation uses a large-scale industrial dataset collected from multi-week ad-serving logs of a prominent e-commerce platform, encompassing over $10^6$ active merchant nodes across heterogeneous marketing categories. To establish an empirical evaluation baseline, we define the \textit{historical Realized Ground Truth} graph $\mathcal{G}_{\text{real}} = (V, \mathcal{E}_{\text{real}})$. 
A directed edge $(v_i, v_j) \in \mathcal{E}_{\text{real}}$ is established if and only if the share of $v_i$'s traffic diverted by $v_j$ within identical auction funnels exceeds a threshold $\tau_e=0.05$ over an observation window (e.g., a one-week tracking period). Although underlying auction collisions are symmetric, this diversion share is inherently asymmetric: $(v_i, v_j)$ may exceed $\tau_e$ while $(v_j, v_i)$ does not.
Consequently, edge existence is often unidirectional (Sec.~\ref{sec:problem}), and reciprocal edges rarely share equal weight, rendering the Top-$K$ set-overlap protocol directionally discriminative.

The evaluation window is strictly isolated from model fitting. Behavioral labels for training the listwise ranker (Sec.~\ref{sec:rerank}) are computed 1 week prior to this window, preventing target leakage.

To quantify how much of this competition represents recently unseen interactions rather than a replay of recent history, we define the \emph{novelty fraction} at cutoff $K$ as the proportion of Top-$K$ realized competitive edges in the evaluation week that never exceeded the diversion threshold $\tau_e$ during the preceding four weeks. This metric rises sharply with $K$, reaching $2.7\%$, $23.6\%$, and $40.5\%$ at $K\in\{10,50,100\}$. This trend demonstrates that while the tightest neighborhoods are dominated by persistent incumbents, over $40\%$ of the Top-100 rivalries are newly emerging edges absent from the recent historical graph.

\subsubsection{Evaluation Metrics}
Let $\mathcal{R}_K(v_i)$ denote the Top-$K$ competitors returned by the final ranking for advertiser $v_i$, and let
$\mathcal{T}(v_i)=\{v_j\mid(v_i,v_j)\in\mathcal{E}_{\text{real}}\}$
denote its empirical realized-rival set. We evaluate over source advertisers with at least one realized competitor and compute advertiser-level Precision@K and Recall@K, averaged across advertisers:
\begin{equation}
P@K =
\frac{1}{|\mathcal{V}_{\text{eval}}|}
\sum_{v_i \in \mathcal{V}_{\text{eval}}}
\frac{|\mathcal{R}_K(v_i) \cap \mathcal{T}(v_i)|}
{|\mathcal{R}_K(v_i)|},
\end{equation}

\begin{equation}
R@K =
\frac{1}{|\mathcal{V}_{\text{eval}}|}
\sum_{v_i \in \mathcal{V}_{\text{eval}}}
\frac{|\mathcal{R}_K(v_i) \cap \mathcal{T}(v_i)|}
{|\mathcal{T}(v_i)|}.
\end{equation}
where $\mathcal{V}_{\text{eval}}=\{v_i \in \mathcal{V} \mid |\mathcal{T}(v_i)|>0\}$. The same $\mathcal{V}_{\text{eval}}$ is used for all evaluated methods. Ranking fidelity is evaluated via NDCG, computed per advertiser and averaged over the same evaluation set, with historical auction co-occurrence counts serving as continuous empirical relevance grades. Because $\mathcal{T}(v_i)$ is a source-specific out-neighbor set dominated by unidirectional edges (Sec.~\ref{sec:problem}), the protocol is inherently directional. To ensure a fair comparison, directional decoding is uniformly applied across all evaluated systems, including both GNN baselines and ablation variants.

Historical ground truth is constrained by selection bias, as interaction logs only document realized auction clashes. Consequently, a latent substitute competitor with high structural rivalry that has not yet entered the same auction pipeline would be counted as a false positive under $\mathcal{G}_{\text{real}}$. To address this limitation and validate such newly uncovered latent rivalries, we establish a statistical alignment protocol in Sec.~\ref{sec:offline_2}.

\subsubsection{GNN Baselines}
\label{sec:external_baselines}
Since existing competitor-mining methods mainly operate at the firm or product level and are not designed for continuously weighted, directed advertiser-level competition (Sec.~\ref{sec:introduction}), we benchmark LACRM against two Graph Neural Network (GNN) references that share LACRM's directed graph, behavioral logs, node universe, and metric protocol. The two GNN references share the same architecture and differ from each other only in the node-level information they consume (full architecture and training details in Appendix~\ref{app:gnn}). \textit{GNN (External)} consumes only behavioral temporal features, quantifying how far graph structure and behavioral dynamics alone can go; \textit{GNN (w/ Qwen Emb.)} uses the same architecture but additionally injects the same pretrained Qwen AR embeddings LACRM uses as static node inputs, isolating whether LLM semantics as node features, without LACRM's asymmetric multi-route retrieval and behavioral alignment, suffice.

\subsubsection{Implementation Details}
The profile generator is \texttt{Qwen3-4B-Instruct}, supervised-fine-tuned on $5{,}000$ \texttt{GPT-5.5} teacher profiles via parameter-efficient LoRA ($r{=}\alpha{=}64$, $4$ epochs, learning rate $10^{-5}$); we subsequently apply the auxiliary GRPO refinement described in Sec.~\ref{sec:profile}. Retrieval runs on the frozen \texttt{Qwen3-Embedding-8B} backbone with no additional embedding-model training. The Dynamic Behavioral Alignment ranker (Sec.~\ref{sec:rerank}) is a LambdaMART ensemble whose listwise NDCG objective, optimized with $\Delta$NDCG-weighted pairwise gradients within each source query group, aggregates cutoffs $K\in\{10,50,100\}$ with weights $\{0.5,0.3,0.2\}$. In production, AR/QR profiles (Sec.~\ref{sec:profile}) are regenerated weekly, reflecting the slower evolution of merchant identity and positioning, while the behavioral features used by Dynamic Behavioral Alignment are recomputed daily to track shorter-term auction dynamics.

\subsection{Quantitative Validation of Realized Market Rivalry}
\label{sec:offline_1}

We first evaluate the system's capacity to retrieve and rank established competitive links empirically manifested in historical logs.

\begin{table*}[!t]
\centering
\caption{Offline retrieval performance (all rows report final post-ranking scores): LACRM against two directed GNN baselines (top block) and an ablation over its progressive representation stages (bottom block), whose barest rung replaces the LLM-synthesized profile with a raw-field semantic-retrieval baseline. Best result per column is in \textbf{bold}.}
\label{tab:offline}
\small
\setlength{\tabcolsep}{8pt}
\begin{tabular}{lccccc}
\toprule
\textbf{Methodology} & \textbf{Precision@100} & \textbf{Recall@100} & \textbf{NDCG@10} & \textbf{NDCG@50} & \textbf{NDCG@100} \\
\midrule
GNN (External) & 0.3442 & 0.5150 & 0.4781 & 0.6334 & 0.6935 \\
GNN (w/ Qwen Emb.) & 0.3628 & 0.6787 & 0.5445 & 0.6395 & 0.7287 \\
\midrule
LACRM (w/o Profile, Multi-view \& Post-training) & 0.3525 & 0.5863 & 0.5069 & 0.6163 & 0.6362 \\
LACRM (w/o Multi-view \& Post-training) & 0.3785 & 0.7156 & 0.5393 & 0.6896 & 0.7349 \\
LACRM (w/o Multi-view) & 0.4157 & 0.7863 & 0.5791 & 0.7207 & 0.7665 \\
LACRM (w/o Post-training) & 0.4107 & 0.7719 & 0.5899 & 0.7275 & 0.7688 \\
LACRM (w/o GRPO) & 0.4641 & 0.8913 & 0.8271 & 0.8906 & 0.8982 \\
\cmidrule(lr){1-6}
\textbf{LACRM (Full)} & \textbf{0.4730} & \textbf{0.9103} & \textbf{0.8437} & \textbf{0.9024} & \textbf{0.9082} \\
\bottomrule
\end{tabular}
\end{table*}

\subsubsection{Comparison with GNN Baselines}
As reported in Table~\ref{tab:offline}, LACRM (Full) outperforms \textit{GNN (External)} on every metric. The behavioral-only GNN remains constrained by the competition graph observed before the evaluation window and therefore has limited access to newly emerging rivals outside that recent graph. Feeding the same Qwen embeddings to the directed GNN (\textit{GNN (w/ Qwen Emb.)}) recovers part of the gap but still trails \textit{LACRM (Full)} substantially in recall. This diagnostic comparison suggests an additional benefit from LACRM's role-separated multi-route retrieval and behavioral reranking beyond static node semantics and directed message passing alone.

\subsubsection{Ablation Study}
\label{sec:ablation_variants}
To isolate the incremental contribution of each progressive representation stage, we anchor the ablation with a profile-free floor, \textit{LACRM (w/o Profile, Multi-view \& Post-training)}, which replaces the LLM-synthesized six-field profile with raw-field embeddings while holding the retrieval and listwise-ranking stack fixed. We then evaluate five additional LACRM variants that successively remove components. \textit{LACRM (w/o Multi-view \& Post-training)} uses single-view QR over an unaligned base model, removing both multi-view expansion and all profile alignment. \textit{LACRM (w/o Multi-view)} keeps post-training but collapses the query representation to a single view with monolithic retrieval. \textit{LACRM (w/o Post-training)} retains the full multi-view routing and listwise ranking layers but uses an unaligned base generator. \textit{LACRM (w/o GRPO)} applies SFT to the generator but omits the auxiliary GRPO refinement, while \textit{LACRM (Full)} retains this refinement in addition to the complete profiling, multi-view retrieval, and behavioral-alignment pipeline.

Table~\ref{tab:offline} presents the offline performance across all model variants, where \textit{LACRM (Full)} consistently achieves the best results across all evaluation metrics.
Post-training gains are dominated by SFT rather than GRPO, indicating that the primary benefit of profile training comes from supervised profile synthesis, while GRPO provides a smaller complementary refinement. One plausible explanation is that SFT establishes the capability to exploit few-shot historical competitor anchors during profile composition (Sec.~\ref{sec:profile}), which transfers directly to inference and can therefore account for much of the gain before the subsequent GRPO refinement.

\subsection{Latent Rivalry Verification} \label{sec:offline_2}

Because latent competitors leave no historical behavioral footprint, $\mathcal{G}_{\text{real}}$ cannot directly confirm them, necessitating a dedicated verification protocol. We assess their legitimacy through two independent and complementary lenses: an ensemble of LLM-as-a-Judge scorers~\cite{gu2024survey} to rate directed competition intensity at scale, and a blind human expert study to guard against shared LLM priors. Both evaluate advertiser pairs on an identical directed competition scale. Together, these complementary evaluations provide evidence that the latent tier contains meaningful competitive relations rather than merely model-generated artifacts.

\subsubsection{LLM-Judge Verification}
We instantiate an asymmetric judge $\mathcal{M}_{\text{judge}}: V \times V \rightarrow [-1, 1]$ that maps each ordered pair $(v_i \rightarrow v_j)$ to a directed competition intensity score $s_{\text{judge}}(v_i \rightarrow v_j)$. This score explicitly quantifies the directional traffic cannibalization and traffic diversion exerted by candidate $v_j$ upon target $v_i$. The score ranges from $+1.0$ (strong traffic diversion with highly homogeneous audiences) to $-1.0$ (unrelated categories with no audience overlap). To reduce systematic bias and generation variance, we sample multiple completions per pair and average the scores $\mathbb{E}[s_{\text{judge}}(v_i \rightarrow v_j)]$. Each judge scores three groups drawn from their respective source populations via reservoir sampling: a positive control $\mathcal{P}_{\text{real}} \subset \mathcal{E}_{\text{real}}$ of $150$ realized rivals, a negative control $\mathcal{P}_{\text{neg}}$ of $150$ random cross-category pairs, and a cohort of $1,500$ newly discovered latent relations $\mathcal{P}_{\text{latent}}$.

We instantiate this evaluation protocol with three LLMs: DeepSeek-V4-Pro~\cite{deepseekai2026deepseekv4highlyefficientmilliontoken}, Doubao-Seed-2.0-pro~\cite{seed2026seed20modelcardintelligence}, and GLM-5.1~\cite{glm5team2026glm5vibecodingagentic}. The final validation relies on ensemble-level agreement rather than any single judge. For each judge, a pair is counted as positive when its score averaged across completions is above zero. The reported $95\%$ confidence intervals are normal-approximation intervals for the cohort mean.

\begin{table*}[!t]
\centering
\caption{Score distributions and stratification metrics across LLM judges. Columns 2--3 and 7--8 report the mean$\pm$std and positive ratio for the realized ($\mathcal{P}_{\text{real}}$) and negative-control ($\mathcal{P}_{\text{neg}}$) cohorts; Columns 4--6 report the mean$\pm$std, $95\%$ confidence interval, and positive ratio for the latent cohort ($\mathcal{P}_{\text{latent}}$). Column 9 reports the Kolmogorov-Smirnov divergence $D_{\text{KS}}$ between $\mathcal{P}_{\text{real}}$ and $\mathcal{P}_{\text{latent}}$. For the Cross-Model Ensemble row, $D_{\text{KS}}$ is the mean of the three per-judge statistics.}
\label{tab:ternary_alignment}
\resizebox{\linewidth}{!}{%
\small
\setlength{\tabcolsep}{5.5pt}
\begin{tabular}{lcccccccc}
\toprule
& \multicolumn{2}{c}{$\mathcal{P}_{\text{real}}$ } & \multicolumn{3}{c}{$\mathcal{P}_{\text{latent}}$}  & \multicolumn{2}{c}{$\mathcal{P}_{\text{neg}}$ } & \multicolumn{1}{c}{\textbf{$\mathcal{P}_{\text{real}}$ vs. $\mathcal{P}_{\text{latent}}$}} \\
\cmidrule(lr){2-3} \cmidrule(lr){4-6} \cmidrule(lr){7-8} \cmidrule(lr){9-9}
\textbf{LLM Base Model} & \textbf{Mean $\pm$ Std} & \textbf{Pos. Ratio} & \textbf{Mean $\pm$ Std} & \textbf{95\% CI} & \textbf{Pos. Ratio} & \textbf{Mean $\pm$ Std} & \textbf{Pos. Ratio} & \textbf{KS ($D_{\text{KS}}$)} \\
\midrule
DeepSeek-V4-Pro & +0.338 $\pm$ 0.458 & 73.3\% & +0.072 $\pm$ 0.510 & [+0.046, +0.098] & 46.0\% & $-0.940 \pm 0.070$ & 0.0\% & 0.273 \\
Doubao-Seed-2.0-pro & +0.557 $\pm$ 0.300 & 94.7\% & +0.341 $\pm$ 0.397 & [+0.321, +0.361] & 78.7\% & $-0.999 \pm 0.008$ & 0.0\% & 0.333 \\
GLM-5.1 & +0.445 $\pm$ 0.491 & 83.3\% & +0.135 $\pm$ 0.581 & [+0.106, +0.164] & 59.3\% & $-0.983 \pm 0.038$ & 0.0\% & 0.287 \\
\midrule
\textbf{Cross-Model Ensemble} & \textbf{+0.447 $\pm$ 0.391} & \textbf{83.8\%} & \textbf{+0.183 $\pm$ 0.462} & \textbf{[+0.160, +0.206]} & \textbf{61.3\%} & \textbf{$-0.974 \pm 0.033$} & \textbf{0.0\%} & \textbf{0.298} \\
\bottomrule
\end{tabular}
}
\end{table*}
\begin{figure*}
    \centering
    \includegraphics[width=0.8\linewidth]{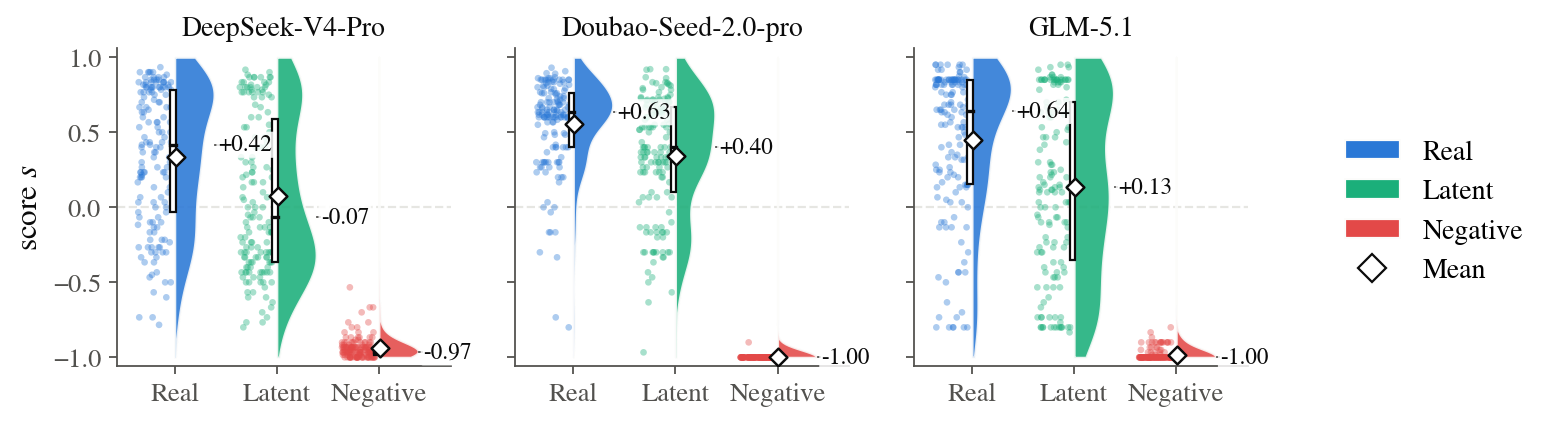}
    \setlength{\abovecaptionskip}{4pt}
    \setlength{\belowcaptionskip}{0pt}
    \caption{Empirical score distributions of the three independent LLM judges. The median score of each group is annotated in the plot.}
    \label{fig:vis}
\end{figure*}

The LLM judges clearly separate the realized and negative controls, with strong cross-judge agreement ($\text{ICC}=0.918$, mean pairwise Spearman $\rho=0.946$; $p<10^{-16}$). As shown in Table~\ref{tab:ternary_alignment} and Figure~\ref{fig:vis}, all three judges consistently rank realized $>$ latent $>$ negative. The latent cohort lies significantly above the negative controls (Mann--Whitney $p<10^{-16}$ for each judge) while remaining distinct from realized rivals (KS $p\approx10^{-9}$), with a $61.3\%$ ensemble positive rate. These results indicate that latent relations form a distinct rivalry stratum rather than noise or a replica of realized competition. We further examine potential shared-LLM bias through independent human verification below.

\subsubsection{Human Verification}\label{sec:human_verification}
To test whether these elevated latent intensities are an artifact of shared LLM priors rather than genuine rivalry, we collect independent human judgments on the latent cohort. Three domain experts independently scored the same $1{,}500$ latent pairs in $\mathcal{P}_{\text{latent}}$ using the unified directed $[-1,+1]$ scale. The evaluation pool also contained $150$ pairs from $\mathcal{P}_{\text{real}}$ and $150$ from $\mathcal{P}_{\text{neg}}$ as embedded validation controls. All control and latent pairs were randomly shuffled and presented without cohort labels or any indication of whether a pair was realized, negative-control, or LACRM-recovered latent, so annotators were blind to sample provenance. Each annotator labeled $100\%$ of the $\mathcal{P}_{\text{real}}$ controls and $0\%$ of the $\mathcal{P}_{\text{neg}}$ controls as positive. On the latent set, all three annotators labeled the large majority of latent pairs as genuine rivalries, with individual positive-label rates of 82.33\%, 85.13\%, and 83.67\%. This substantial agreement is supported by an average-measures reliability of $\text{ICC}(2,3)=0.846$ and a single-measure reliability of $\text{ICC}(2,1)=0.647$.

\subsection{Coverage and Generalization Analysis}

\begin{table}[t]
\centering
\small
\setlength{\tabcolsep}{5pt}
\caption{Per-view activation: the fraction of Top-$K$ retrieved candidates hit by each competitive route (sorted by Top-10).}
\label{tab:view_hit}
\begin{tabular}{@{}p{3.5cm}ccc@{}}
\toprule
\textbf{Competitive Route} & \textbf{Top-10} & \textbf{Top-50} & \textbf{Top-100} \\
\midrule
Intra-Brand Cross-Channel & 87.8\% & 82.3\% & 79.7\% \\
Direct Homogeneous Rivalry & 85.2\% & 83.9\% & 84.9\% \\
Intra-Brand Structural & 73.9\% & 68.0\% & 66.2\% \\
Temporal Campaign & 37.5\% & 34.0\% & 32.2\% \\
Cross-Category Functional & 18.6\% & 17.5\% & 17.0\% \\
\midrule
Avg.\ views hit / candidate & 3.03 & 2.86 & 2.80 \\
\bottomrule
\end{tabular}
\end{table}

\subsubsection{Per-View Activation}
We compute the activation rate for each view, defined as the fraction of Top-$K$ candidates it retrieves across $K\in\{10,50,100\}$ (Table~\ref{tab:view_hit}). 

View activation is highly non-uniform: dense primary routes (\textit{Intra-Brand Cross-Channel Friction} and \textit{Direct Homogeneous Rivalry}) account for a large share of top-ranked candidates ($80$--$88\%$), while specialized sparse routes (\textit{Cross-Category Functional Substitution} and \textit{Temporal Campaign Clashes}) expand coverage to less frequent competition patterns ($17$--$37\%$).

\subsubsection{Generalization to Emerging Competition Edges}
The \textit{novelty fraction} indicates that a substantial fraction of evaluation-week competition consists of edges absent from recent history. Consequently, a model that merely memorizes the recent historical graph struggles to generalize to these emerging edges. We therefore isolate performance on the emerging-edge subset: for each cutoff $K$, we restrict the evaluation-week ground truth to its Top-$K$ realized edges, keep only those that never exceeded $\tau_e$ in any of the preceding four weeks, and compare the model's recall on this emerging subset against its recall on the complementary historical edges that were already active during those four weeks.

LACRM achieves emerging-edge Recall of $8.6\%$, $77.9\%$, and $86.0\%$ at $K\in\{10,50,100\}$, respectively, showing that most edges absent from the recent four-week competition graph are recovered once the candidate neighborhood is moderately expanded. The recall gap between historical and emerging edges correspondingly shrinks from $66.80$ percentage points (pp) at $K{=}10$ to $16.39$ pp and $8.41$ pp at $K{=}50$ and $K{=}100$, respectively. Because these edges are defined as inactive during the preceding four weeks, the results show that LACRM is not restricted to currently active historical edges and retains high recall on recently unseen competition at larger cutoffs.

\subsection{Online Experiments}

We conduct two online A/B experiments on a large-scale e-commerce advertising platform to evaluate LACRM in a production environment. Both experiments follow the same experimental protocol: a one-week pre-period for baseline calibration, a short washout gap to exclude ramp-up effects, and a $14$-day treatment window. The first experiment examines whether LACRM-derived competition relationships improve marketplace budget utilization, while the second evaluates their effectiveness in supporting competition-aware advertiser incentives.

Naive unit-level randomization can induce cross-arm interference in auction-based advertising markets, because treatment and control advertisers may compete in the same auctions, causing one advertiser's assignment to affect the outcomes of others; budget and throttling constraints can further distort treatment-effect estimates~\cite{basse2016randomization}. We therefore use cluster-level randomization~\cite{ugander2013graphcluster,eckles2017networks} in both experiments and treat each advertiser cluster as an atomic assignment unit.

\paragraph{Experimental Design.}
For each experiment, eligible advertisers are drawn from a precomputed non-overlapping high-cohesion partition derived from LACRM's inferred competition graph before treatment assignment, grouping advertisers connected by strong mutual competitive pressure into the same cluster. The resulting clusters, rather than separately constructed arm-specific partitions, form the common randomization pool. We select $G=200$ eligible clusters and randomly assign them at a $50{:}50$ ratio to treatment ($100$) and control ($100$), covering approximately $100{,}000$ advertisers per experiment; the selected partition is held fixed across arms. This design reduces cross-arm interference along competition relationships represented by LACRM, although residual interference may remain for competitive relationships not captured by the inferred graph, including relationships identified only by the rule-based \textbf{Base} baseline. Treatment effects are estimated with a difference-in-differences estimator relative to each arm's pre-period outcomes. Standard errors are clustered at the advertiser-cluster level, matching the unit of randomization. We additionally assess significance using randomization-inference $p$-values from $10{,}000$ cluster-level treatment assignments that preserve the $100{:}100$ split, with a fixed seed. This protocol is shared by both online experiments below.

\subsubsection{Competition-Aware Bidding Experiment}

Both conditions use the same reinforcement-learning bidding backbone and differ only in the competition signal supplied to the bidding coordination mechanism. In the control condition (\textbf{Base}), competition relationships come from the platform's rule-based industry baseline, which relies on merchant attributes such as fine-grained product categories, merchant tiers, and store operation models. In the treatment condition (\textbf{LACRM}), each advertiser's own bidding policy instead uses the competition relationships inferred by our framework.

Under this design, the LACRM-guided condition yields a $1.73\%$ increase in ad revenue and a $0.91$ percentage-point improvement in the daily budget-target fulfillment rate. Both effects are statistically significant under the cluster-level inference procedure, with randomization inference yielding the same significance conclusion. These results suggest that LACRM-derived competition relationships provide a more effective competition signal for auction-level bidding coordination and advertiser budget utilization than the rule-based baseline.

\subsubsection{Competition-Aware Incentive Experiment}

We further conduct a complementary experiment under the same design to evaluate whether LACRM-derived competition relationships can support more precise advertiser incentives. The control arm follows the platform's existing bid- and budget-based incentive policy, whereas the treatment arm augments this mechanism with LACRM-derived competition signals. Each advertising unit is evaluated relative to its competitors operating under comparable market conditions, such as similar price bands, so that incentive allocation can account for the advertiser's relative position within its competitive environment.

Following common industry practice, eligible advertisers receive conditional promotional ad credits designed to encourage active bid expression and sustained budget commitment. The total credit budget is held fixed and evenly split across treatment and control, so the two arms receive the same aggregate credit budget and differ only in how those credits are allocated. This rules out a larger aggregate subsidy budget as the source of the reported treatment effect. To protect commercially sensitive information, we omit the specific eligibility thresholds, credit amounts, and delivery rules.

Under the shared difference-in-differences design, the LACRM-guided incentive mechanism yields a $1.14\%$ increase in ad revenue relative to the existing incentive policy. The effect is statistically significant under the same cluster-level inference procedure ($p<0.01$). This result suggests that LACRM-derived competition signals can improve incentive precision and encourage advertisers to express their valuations more actively through bids and budget commitments.

\section{Related Work}

Online advertising and sponsored search build on auction and mechanism design~\cite{myerson1981optimalauctiondesign,VARIAN20071163,edelman2007gsp}, with prior work also studying competitive externalities and strategic spillovers in targeting and allocation~\cite{bergemann2011targeting,jehiel1996strategicnonparticipation,ghosh2008externalities,sayedi2014competitivepoaching}. Outside computational advertising, asymmetric competition has long been studied in marketing science: Carpenter et al.~\cite{carpenter1988modeling} characterize asymmetric competitive effects across brands, while Netzer et al.~\cite{netzer2012mine} recover brand-level market structure from user-generated text. The auction-derived traffic-diversion signal measures directional traffic substitution between advertisers, paralleling the diversion-ratio formulation used in antitrust analysis~\cite{farrell2010antitrust}. These studies establish that asymmetric competition and data-driven market-structure recovery are not new concepts. Our focus is their advertiser-level operationalization: recovering directed, continuously weighted competition through semantic retrieval aligned with temporal auction behavior for downstream advertising optimization.

Competition mining has mainly been studied at the product or firm level. Behavioral methods infer substitute or complement relations from co-purchase and basket data~\cite{mcauley2015substitutable,chen2020studying}, while graph-based approaches construct company networks or retrieve competitors from financial and inter-firm knowledge graphs~\cite{cao2023companykg,ding2024jpec,intercorprel2025}. Such methods primarily capture product substitutability, firm similarity, or observed graph relations, and do not directly model continuously varying directional competitive pressure among advertisers.

LLM-based recommendation uses language models to enrich sparse behavioral signals and external knowledge~\cite{xi2023kar,llmrec2024,hou2024llm4rec,wang2024recmind}, while text encoders such as E5, BGE, and Qwen3-Embedding provide strong semantic retrieval capabilities~\cite{wang2022e5,xiao2023bge,qwen3embedding}. However, these methods target general semantic relevance or user--item ranking rather than directed advertiser competition. LACRM instead combines role-separated semantic retrieval with temporal auction-behavior alignment to recover a directed, weighted advertiser competition graph.

\section{Conclusion}
In this paper, we focus on the problem of mining directed advertiser competition relationships in e-commerce advertising and formulate the underlying market structure as a directed, weighted graph. We propose LACRM, an LLM-guided framework that integrates role-separated advertiser profiling, asymmetric multi-route retrieval, and listwise behavioral alignment. This design enables LACRM to capture heterogeneous competition patterns, distinguish directional competitive pressure, and discover latent rivals beyond historical auction interactions.
On large-scale industrial advertising data, LACRM consistently outperforms the evaluated directed GNN baselines and ablated variants in realized-rival retrieval and ranking. Independent LLM and human assessments further support the competitive relevance of its recovered latent rivals.
In production experiments, LACRM-derived competition signals yield a 1.73\% increase in ad revenue and a 0.91 percentage-point improvement in budget-target fulfillment for bidding, as well as a 1.14\% increase in ad revenue for advertiser incentives, supporting their practical value in advertiser-facing optimization.

\newpage
\appendix

\section{Query-View and Behavioral-Feature Specifications}
\label{app:views}

We detail the field subsets and competitive boundaries of the five query routes (Sec.~\ref{sec:asymmetric_embedding}) and the per-group construction of the behavioral profile consumed by the ranker (Sec.~\ref{sec:rerank}). Each route operates over a designated field subsystem $\mathcal{F}^{(\omega)} \subseteq \mathcal{F}$.

\vspace{-0.25cm}
\subsection{Five-View Field Assignment} \textbf{Direct Homogeneous Rivalry} ($f_1, f_2, f_5$) is the high-recall structural anchor, isolating pairwise comparisons between interchangeable commodities. The two brand-centric routes share the field subset ($f_1, f_6$) but split on the cross-L3 vs.\ same-L3 boundary. Horizontal expansion of one brand across divergent L3 categories (e.g., an electronics brand entering smart wearables) corresponds to \textbf{Intra-Brand Structural Expansion}. In contrast, an identical brand and category competing across divergent channels (e.g., first-party supply vs.\ an official flagship store) forms \textbf{Intra-Brand Cross-Channel Friction}. The two routes are enforced as mutually exclusive by a prompt-level boundary gate. \textbf{Cross-Category Functional Substitution} ($f_2, f_3, f_4, f_5$) masks the category dimension to surface implicit substitutes competing for the same budget or usage scenario (e.g., a smartwatch versus a luxury timepiece as a gift), serving as the primary route for latent, long-tail threats. \textbf{Temporal Campaign Clashes} ($f_2, f_4$) captures time-gated dynamics active exclusively during promotional windows. When festival contexts are absent, the corresponding QR is marked as \emph{inactive} and the ANN query for this route is skipped. Consequently, no candidates are retrieved from this view, and its entries in the downstream hit-indicator vector $\mathbf{m}_{ij}$ (Sec.~\ref{sec:asymmetric_embedding}) remain zero. Otherwise, active requests are served directly through the shared index without requiring a dedicated online branch.

\vspace{-0.25cm}
\subsection{Behavioral Feature Construction}
\label{app:features}
The six groups of the behavioral profile $\tilde\bt_i$ (Sec.~\ref{sec:rerank}) are instantiated as: (1) \textbf{Traffic Scale}: log-scaled impressions, clicks, and GMV over rolling 1/7/14-day windows; (2) \textbf{Conversion Efficiency}: short-window CTR, CVR, and average order value; (3) \textbf{Volumetric Dynamics}: market-share trajectories, tier-wise Z-scores, coefficients of variation, and traffic-saturation metrics such as zero-exposure rates; (4) \textbf{Ecosystem Rhythm \& Lifecycle}: weekly Fourier periodicities, active ad-serving days in the current quarter, and platform tenure; (5) \textbf{Behavioral Quality}: funnel indicators including add-to-cart ratios, promotional-discount depth, and revenue per page view; and (6) \textbf{Intra-Category Quantiles \& Trends}: category-bounded GMV percentiles and multi-lag autocorrelations tracking competitive displacement within the same vertical.

\section{GNN Reference Implementation Details}
\label{app:gnn}

The two GNN references in Table~\ref{tab:offline} are diagnostic baselines that share LACRM's directed graph, behavioral logs, node universe, directed ground truth, and metric protocol (Top-$K$ retrieval scored against $\mathcal{G}_{\text{real}}$), differing only in the node-level inputs they consume (detailed below). Their comparison with LACRM provides a diagnostic view of the additional benefit of the full LLM-guided design.

\paragraph{Competition Graph.}
Nodes are advertisers $v_i \in V$; edges are constructed exclusively from realized competition records observed strictly prior to the label window, thereby precluding temporal leakage. Each realized directed rivalry contributes a forward competition edge $v_i \!\to\! v_j$ together with a distinctly-typed reverse edge $v_j \!\to\! v_i$, so the encoder can propagate incoming and outgoing competitive pressure through separate relation channels. Daily observations of the same directed pair are aggregated into a single edge weight that fuses competitive intensity with temporal decay,
\begin{equation}
  w_{ij} = \sum_{d \in \mathcal{H}} \log\!\bigl(1 + I_{ij}^{(d)}\bigr)\,\exp\!\Bigl(-\tfrac{\text{age}(d)}{\tau_d}\Bigr),
\end{equation}
where $I_{ij}^{(d)}$ is the realized competition intensity on historical day $d$, $\text{age}(d)$ is the day's distance to the graph cutoff, and $\tau_d$ is a decay constant set to 14 days.

\vspace{-0.25cm}
\paragraph{Node Features.}
The two references share an identical encoder, decoder, and training objective, and differ only in node inputs. \textbf{GNN (External)} carries no LLM-derived signal: it drops LACRM's generative AR/QR profiles, the instruction-conditioned semantic score $s_{\text{emb}}$, and the semantic alignment objective, so each node holds only the behavioral temporal feature vector of Sec.~\ref{sec:rerank}. \textbf{GNN (w/ Qwen Emb.)} instead sets each node's input to the same pretrained Qwen embedding of the advertiser AR profile used throughout LACRM. Contrasting the two provides a diagnostic estimate of the value of LLM semantics as static node features, while their gap to LACRM suggests the additional benefit of asymmetric multi-route retrieval and behavioral alignment.
\vspace{-0.25cm}
\paragraph{Encoder.}
We use a two-layer directed, relation-aware GraphSAGE~\cite{hamilton2017inductive} encoder. Within each layer, incoming and outgoing neighborhoods are aggregated with independent per-relation transformation parameters, and messages are weighted by $w_{ij}$ under mean aggregation; the aggregated neighborhood is concatenated with a self (root) transform and passed through an MLP with a residual connection and layer normalization. Separate in-/out-edge parameters let the two-hop states encode directional, relation-specific context rather than a symmetric neighborhood average. Hidden, output, and role dimensions are all $128$ (node/edge feature encoders project to $64$), with dropout $0.1$ and \textsc{gelu}/\textsc{elu} layer activations; the two layers sample edge-weight- and recency-weighted fan-outs of $40$ and $20$ neighbors.

\vspace{-0.25cm}
\paragraph{Asymmetric Decoder.}
Edge scores are produced by an asymmetric decoder that maps each endpoint through distinct source- and target-role heads, combines them with a low-rank bilinear term, and passes the interaction (bilinear score, elementwise product, and difference of the two role vectors) through an edge MLP. Because the source and target projections are untied, the decoder defines a directional scoring function, and $s(v_i\!\to\!v_j)$ is not constrained to equal $s(v_j\!\to\!v_i)$, matching the asymmetry of the competition relation.

\vspace{-0.25cm}
\paragraph{Training Objective.}
Both references are trained with the same listwise LambdaRank objective as LACRM (Sec.~\ref{sec:rerank}). We optimize with AdamW~\cite{loshchilov2019decoupled} (learning rate $3\times10^{-4}$ with linear warmup and decay). To rule out under-fitting of these external references, both are trained to convergence rather than to a fixed budget: we monitor validation NDCG on a held-out split and apply early stopping (patience $3$, minimum improvement $10^{-4}$, evaluated from epoch $5$ up to a $10$-epoch cap); all runs stopped early before reaching the cap, and the reported metrics are taken at this converged checkpoint.

\bibliographystyle{ACM-Reference-Format}
\bibliography{sample-base}

\end{document}